\documentclass{iopjournal}

\usepackage[utf8]{inputenc} % allow utf-8 input
\usepackage[T1]{fontenc}    % use 8-bit T1 fonts
\usepackage{subcaption}     % side-by-side subfigures
\usepackage{url}            % simple URL typesetting
\usepackage{booktabs}       % professional-quality tables
\usepackage{amsmath}        % align, equation helpers
\usepackage{amsfonts}       % blackboard math symbols
\usepackage{nicefrac}       % compact symbols for 1/2, etc.
\usepackage{microtype}      % microtypography
\usepackage{makecell}
\usepackage{pifont}
\usepackage{natbib}         % author-year (Harvard) citations

\newcommand{\healpix}{\textsc{HEALPix}}
\newcommand{\mathd}{\ensuremath{\mathrm{d}}}
\newcommand{\imagunit}{\ensuremath{\textrm{i}\mkern1mu}}
\newcommand{\subsubsect}[1]{\smallskip \noindent \textbf{#1:}}
\newcommand{\Ngam}{N_{\gamma}}
\newcommand{\Npix}{N_{\mathrm{pix}}}

\renewcommand{\articletype}[1]{{\vspace*{-8mm}\noindent \Large \sf Machine Learning: Science and Technology}

\vspace*{8mm} \noindent\reversemarginpar
\marginpar{\vspace{-3mm} {\color{gray}\hrule} \ \\ Crossmark\\ {\color{gray}\hrule} \ \\ \tiny {\sf RECEIVED} {\small \\ 1 October 2026}\\ \\ {\sf REVISED} {\small \\ dd Month yyyy}}{\scriptsize \sf{\bfseries \MakeUppercase{#1}}}}

\begin{document}

\articletype{Paper}

\title{Learning without relearning physical symmetry: an SO(3)-equivariant convolutional network (CNN) on the sphere}

% TODO: add ORCID iDs with \orcid{0000-0000-0000-0000} after each name,
% and confirm the corresponding author.
\author{
Joshua Williamson$^{1,*}$,
Niall Jeffrey$^{2,1}$,
Lorne Whiteway$^{1}$
}

\affil{$^1$Department of Physics \& Astronomy, University College London, Gower Street, London, WC1E 6BT, UK}

\affil{$^2$Department of Physics \& King's Institute for Artificial Intelligence, King's College London, Strand, London WC2R 2LS, United Kingdom}

\affil{$^*$Author to whom any correspondence should be addressed.}

\email{joshua.williamson.21@ucl.ac.uk}


\begin{abstract}
Many of the most important datasets in the physical sciences live natively on the sphere, and these data often carry no preferred position or orientation. Using methods that do not reflect this symmetry leads to far higher simulation requirements, which can be catastrophic for scientific analyses (forcing reliance on cheap, imperfect models that bias results). We present a novel spherical implementation of a gauge equivariant mesh architecture that is specifically designed for the Hierarchical Equal Area isoLatitude Pixelization (\healpix{}) of the sphere. The convolution is equivariant under $\mathrm{SO}(3)$, the group of three-dimensional rotations, so rotating the input map rotates the output in the same way. It is built from steerable filters (i.e. filters that can be rotated to any orientation exactly, without resampling) and spin-weighted features. The architecture supplies the exact symmetry, so the network does not spend training data learning it. The convolution operations are constructed to give provably exact equivariance, given a band-limit. Our full end-to-end CNNs can achieve SO(3)-invariance error at per-cent-level and lower. On benchmark tasks, our network gives the most accurate dense gradient regression of Gaussian random fields, with an error 55 per cent lower (a factor of 2.2) than the best alternative, and is competitive in cosmic microwave background foreground removal and denoising, a task in which anisotropic instrumental noise breaks the rotational symmetry. Gains in simulation efficiency matter most in simulation-based inference (SBI), where high-fidelity simulations are expensive and neural compressors require large training sets; a fixed simulation budget therefore forces analyses onto cheaper, less accurate simulators. In weak lensing SBI at matched simulation count, our network yields the tightest posteriors on $\Omega_{\rm m}$ and $S_8$, improving the mean Figure of Merit by factors of 1.45 and 2.92 over graph-based and axisymmetric discrete-continuous spherical convolutions, respectively. This approach enables new convolutional neural network analyses of processes and data on the sphere, including generative and inference tasks.
\end{abstract}

% ============================================================
% Main text
% ============================================================

\vspace{2cm}

\section{Introduction}

Many of the most important datasets in the physical sciences live natively on the sphere. The cosmic microwave background (CMB), a probe that is central to our standard model of cosmology, is measured over the full celestial sphere \citep{planck2020}; weak gravitational lensing surveys map the large-scale structure of the Universe \citep{jeffrey2024dark}; and weather forecasting and climate science operate on fields covering the Earth's sphere \citep{bonev2023}.

Increasingly, extracting information from these high-dimensional datasets relies on large suites of numerical simulations, most directly in simulation-based inference (SBI). High-fidelity simulations are often extremely expensive, however, forcing practical analyses to rely on cheaper, imperfect models in order to generate sufficient training data. Any mismatch between these models and reality propagates directly into the inferred posteriors. Reducing the number of simulations required can reduce this dependence on imperfect models. The more efficient the training, the less imperfect the models, as the simulation budget is spent on fidelity rather than simulation count. We provide one route to such a reduction: by encoding known physical structure as an inductive bias, the learning problem requires fewer examples. For spherical data that structure is symmetry. These data often have no preferred position or orientation, motivating architectures that are equivariant under $\mathrm{SO}(3)$, the group of three-dimensional rotations.

Planar convolutional neural networks (CNNs) are poorly matched to this geometry, as any projection of the sphere onto the plane distorts areas and distances, and translational weight sharing corresponds to no symmetry of the sphere. A network without rotation-equivariant filters must learn the same feature at every position and orientation, seeing it repeated across the training data through augmentation or natural variability. SO(3)-equivariant representations remove this redundancy, sharply reducing training data requirements. When every training example is an expensive simulation, exploiting this symmetry offers a direct route to doing more science with fewer simulations. For SBI this matters twice over, since the simulation suite is both the training set and the sole source of the likelihood. Any reduction in simulation fidelity therefore enters the analysis as model misspecification, biasing the inferred posteriors.

Convolutions that are both rotationally equivariant and scalable have proven difficult to construct (Table~\ref{tab:method_comparison}). Group-based spherical CNNs built on Wigner transforms \citep{cohen2018, esteves2018, kondor2018, cobb2021} are exactly equivariant but costly, and their scalable spin-weighted descendants \citep{esteves2020, esteves2023} remain bound to spectral transforms on full sky maps. Spatially localized alternatives, such as hybrid discrete-continuous (DISCO) convolutions \citep{ocampo2023} and graph constructions on the Hierarchical Equal Area isoLatitude Pixelization (\healpix{}) \citep{perraudin2019, defferrard2020}, are only approximately rotationally equivariant at finite resolution.

CNNs have been used already in SBI on probes of weak lensing and CMB maps (\citealt{jeffrey2024dark}; \citealt{williamson2026}; \citealt{thomsen2025, wolz2023}), and global weather forecasting \citep{hu2025smgno}, yet few have attempted architectures that enforce strict $\mathrm{SO}(3)$ equivariance. In this paper we demonstrate that exact $\mathrm{SO}(3)$ equivariance can be carried onto the \healpix{} grid at practical cost, and that in a realistic weak lensing SBI task it yields tighter posteriors from the same simulation suite compared to planar, graph-based, and axisymmetric DISCO alternatives.

\begin{figure}[t]
  \centering
  \begin{subfigure}[t]{0.49\linewidth}
    \centering
    \includegraphics[width=\linewidth]{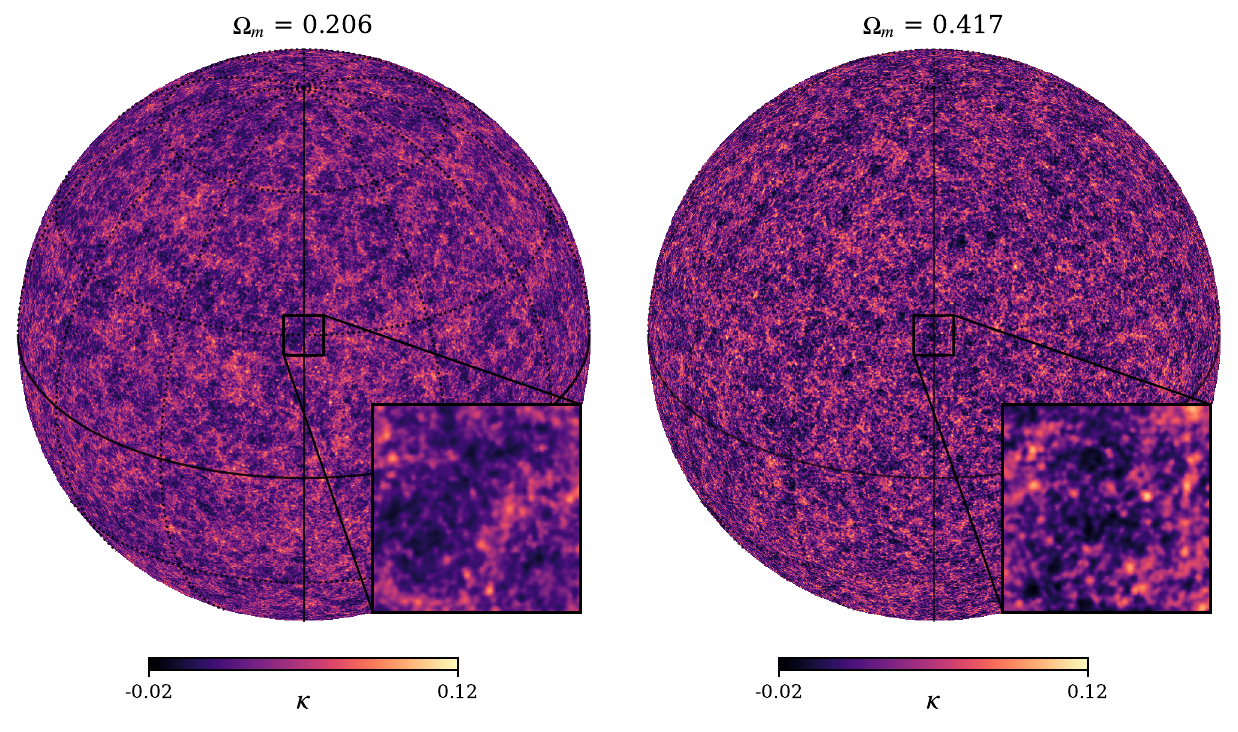}
    \caption{Weak-lensing convergence $\kappa$ for two values of $\Omega_{\rm m}$.}
    \label{fig:kappa}
  \end{subfigure}
  \hfill
  \begin{subfigure}[t]{0.49\linewidth}
    \centering
    \includegraphics[width=\linewidth]{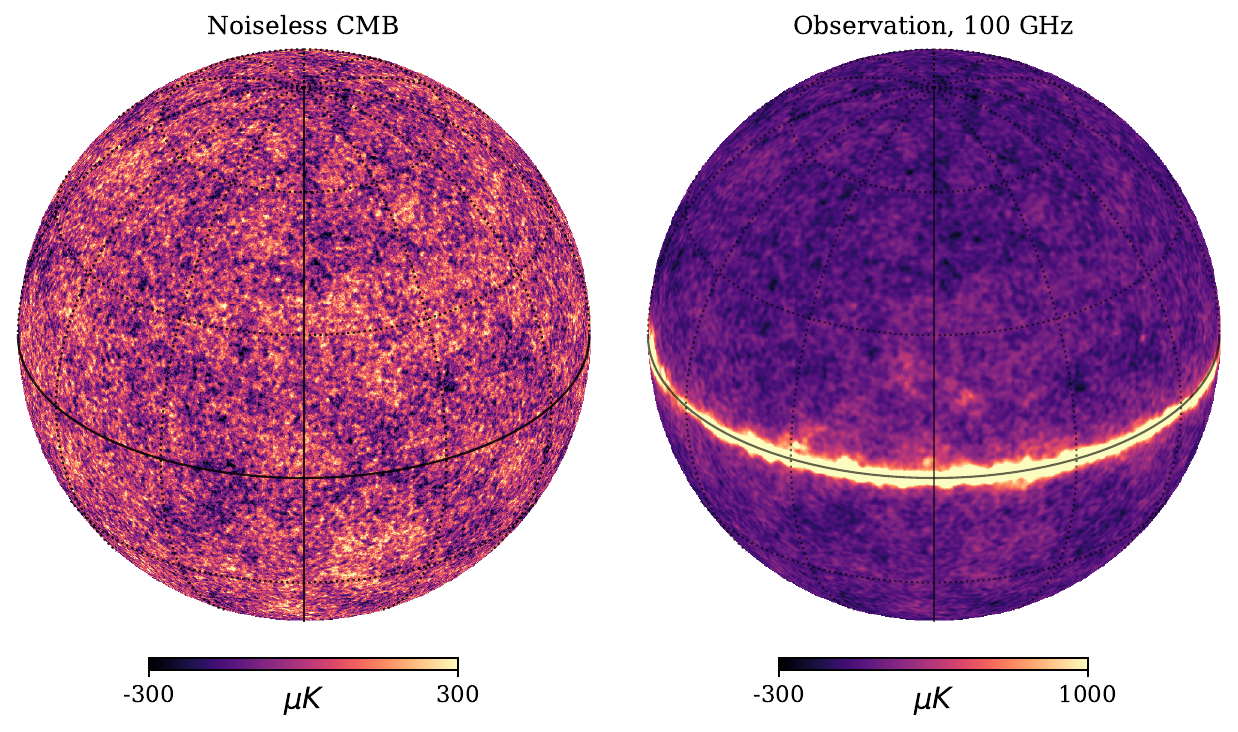}
    \caption{Noiseless CMB map and a corresponding 100\,GHz mock observation.}
    \label{fig:cmb}
  \end{subfigure}
  \caption{Spherical data from cosmology for the benchmark tasks.}
  \label{fig:spherical_maps}
\end{figure}

\section{SO(3) convolution method}\label{sec:so3}

Rotating a directional filter on the sphere changes its response, so a rotation-equivariant network must record the response at every orientation as well as at every position. Its feature maps are therefore functions on the rotation group $\mathrm{SO}(3)$, and its layers are group convolutions \citep{cohen2016, cohen2018}. We evaluate these convolutions by quadrature with the filter kept continuous \citep{ocampo2023}, expand the filter in circular harmonics \citep{freeman1991, worrall2017}, and relate the local frames at neighbouring pixels by parallel transport \citep{dehaan2021}. None of these components are new; our contribution is their combination on the \healpix{} grid in a form that is exactly, rather than approximately, equivariant. Derivations beyond those given here can be found in these papers and references therein.

\subsection{Rotations as positions and orientations}\label{sec:so3:rot}

Let $\mathbb{S}^2$ be the unit sphere, with points labelled by colatitude $\theta \in [0,\pi]$ and longitude $\phi \in [0,2\pi)$, and let SO(3) be the group of rotations of three-dimensional space. In the $zyz$ Euler convention every rotation can be written as $R = Z(\phi) Y(\theta) Z(\gamma)$, where $Z(\cdot)$ and $Y(\cdot)$ denote rotations about the $z$ and $y$ axes. The first two factors carry the north pole $\hat{z}$ to the point $x = (\theta,\phi)$, and $\gamma \in [0,2\pi)$ is a rotation about the axis through $x$, so a rotation specifies a position $x$ on the sphere together with an orientation $\gamma$ at that position. We write $g = (x,\gamma)$ for the rotation with these coordinates and call $g$ a pose.

The angle $\gamma$ is measured from a reference direction in the tangent plane at $x$, which the Euler convention takes to be the meridian direction, i.e. local north. No such direction can be chosen continuously over the whole sphere, and this one is undefined at the two poles. This suffices for integration and, since no \healpix{} pixel centre lies at a pole, for every sample point that we use. The arbitrariness of north is the origin of the transport angle of Section~\ref{sec:so3:pose}.

Integration over the group uses the Haar measure $\mu$, the unique measure up to normalization that is invariant under rotation, $\mu(QS) = \mu(S)$ for every rotation $Q$ and every measurable set $S$. In Euler coordinates $\mathd \mu = \sin\theta \, \mathd \theta \, \mathd \phi \, \mathd \gamma$, the product of the area element $\mathd \mu_{\mathbb{S}^2} = \sin\theta \, \mathd \theta \, \mathd \phi$ on the sphere and the uniform measure $\mathd \gamma$ on the circle of orientations. A signal on the group is a function $F(x,\gamma)$ and a signal on the sphere a function $f(x)$; channel indices are suppressed.

\subsection{Convolution on the rotation group}\label{sec:so3:conv}

Given a signal $F$ on SO(3) and a filter $\psi$, which is also a function on SO(3), their group convolution is
\begin{equation}
H(g) \equiv (F \star \psi)(g) = \int_{{\rm SO}(3)} F(h) \, \psi(g^{-1} h) \, \mathd \mu(h) .
\label{eq:conv}
\end{equation}
Here $g$ is the pose at which the output $H$ is evaluated, $h$ is the integration variable running over all input poses, and $g^{-1}h$ is the pose of $h$ relative to $g$, i.e. the rotation that carries the frame $g$ to the frame $h$. Evaluating the filter on the relative pose is what makes equation~(\ref{eq:conv}) equivariant. For any rotation $Q$ define the rotated signal $(Q \cdot F)(h) = F(Q^{-1}h)$; substituting $h = Qh'$ and using the invariance of $\mu$,
\begin{equation}
((Q \cdot F) \star \psi)(g) = \int F(h') \, \psi(g^{-1} Q h') \, \mathd \mu(h') 
= (F \star \psi)(Q^{-1} g) = (Q \cdot H)(g) ,
\label{eq:equiv}
\end{equation}
so rotating the input rotates the output by the same rotation \citep{cohen2016}. This is \textit{equivariance}, the desired property.

Equation~(\ref{eq:conv}) maps signals on SO(3) to signals on SO(3) and is the form of every layerof the convolutional neural network, except for the first layer. The first layer receives a signal $f$ on the sphere, and its lift to the group is
\begin{equation}
H(g) = \int_{\mathbb{S}^2} f(x) \, \psi(g^{-1} x) \, \mathd \mu_{\mathbb{S}^2}(x) ,
\label{eq:lift}
\end{equation}
where $\psi$ is now a function on the sphere and $g^{-1}x$ is the point $x$ expressed in the frame $g$ \citep{cohen2018, esteves2018}. The output depends on the orientation of $g$ because the filter is anisotropic, and equivariance follows as in equation~(\ref{eq:equiv}). 

\begin{figure}[t]
  \centering
  \includegraphics[width=\linewidth]{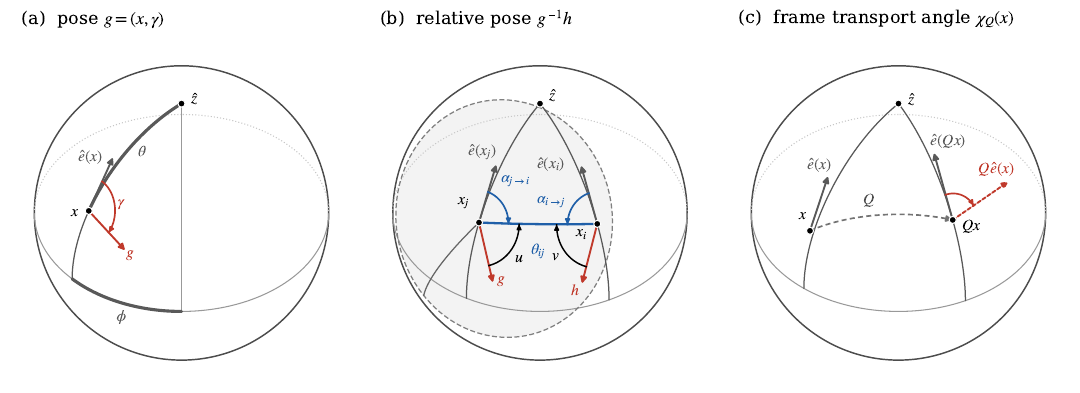}
  \caption{Coordinates of the SO(3) convolution: a pose, the relative pose of two points, and the transport angle.}
  \label{fig:so3_coords}
\end{figure}

\subsection{Relative pose in coordinates}\label{sec:so3:pose}

We refer the reader to Fig.~\ref{fig:so3_coords} to aid geometric understanding. To evaluate $\psi(g^{-1}h)$ we need coordinates for the relative pose. Let $g = (x_j,\gamma)$ and $h = (x_i,\gamma')$. The separation $\theta_{ij} \in [0,\pi]$ is the length of the shorter great-circle arc joining $x_i$ and $x_j$. This is unique unless the points are antipodal, though our filters have a cutoff $\theta_{\rm c} < \pi$ meaning that this never arises.

Writing $\hat{e}(x)$ for the unit tangent vector at $x$ pointing along the meridian towards the north pole, the bearing $\alpha_{j \to i} \in [0,2\pi)$ is the angle at $x_j$ from $\hat{e}(x_j)$ to the direction of the arc towards $x_i$. The bearing $\alpha_{i \to j}$ is defined in the same way at $x_i$. These are two are independent quantities.

The bearings depend on the choice of north, but the differences
\begin{equation}
u = \alpha_{j \to i} - \gamma , \qquad v = \alpha_{i \to j} - \gamma'
\label{eq:uv}
\end{equation}
do not: $u$ is the direction of $x_i$ measured from the reference direction of the frame $g$, and $v$ the direction of $x_j$ measured from that of the frame $h$. Since $g^{-1}h$ moves the pole to colatitude $\theta_{ij}$ and azimuth $u$ in the frame $g$, with the remaining frame rotation given by $v$ (up to an offset absorbed into the filter), $\psi(g^{-1}h)$ is a function $\psi(\theta_{ij},u,v)$ of these three numbers.

A rotation $Q$ carries $x$ to $Qx$ and carries $\hat{e}(x)$ to a tangent vector at $Qx$ which in general is not $\hat{e}(Qx)$; the two differ by an angle $\chi_Q(x)$, the transport angle \citep{cohen2019, dehaan2021}. In the coordinates of Section~\ref{sec:so3:rot} the rotation therefore acts as
\begin{equation}
Q \cdot (x,\gamma) = (Qx, \, \gamma + \chi_Q(x)) ,
\label{eq:action}
\end{equation}
and since every bearing at $x$ is measured from the same $\hat{e}(x)$, it shifts by the same $\chi_Q(x)$ as $\gamma$ does. The differences in equation~(\ref{eq:uv}) are unchanged, as is $\theta_{ij}$, which is why the integrand of equation~(\ref{eq:conv}) is invariant and the convolution equivariant. 
\subsection{Filter basis}\label{sec:so3:filter}

The filter is a function on the cap $\theta \le \theta_{\rm c}$, periodic in $u$ and in $v$, which we expand as
\begin{equation}
\psi(\theta,u,v) = \sum_{p=1}^{P} \sum_{m=-M}^{M} \sum_{n=-N}^{N} c_{pmn} \, R_p(\theta) \, e^{\imagunit m u} \, e^{\imagunit n v} .
\label{eq:filter}
\end{equation}
The $R_p$, $p = 1, \dots, P$, are fixed radial profiles vanishing for $\theta > \theta_{\rm c}$ (we use Zernike radial polynomials), $m$ and $n$ are the angular orders in $u$ and $v$, truncated at $M$ and $N$, and the complex coefficients $c_{pmn}$ are the only learnable parameters; smoothness at $\theta = 0$ requires $R_p(0) = 0$ whenever $m \ne 0$. These are the circular harmonic, or steerable, filters of planar rotation-equivariant networks \citep{freeman1991, worrall2017} transferred to the sphere, adopted in place of the interpolated filters of \citet{ocampo2023} because, by equation~(\ref{eq:uv}), $e^{\imagunit m u} = e^{-\imagunit m \gamma} e^{\imagunit m \alpha_{j \to i}}$, so changing the output orientation $\gamma$ multiplies each term by a phase, and the filter is steered to any orientation exactly, without resampling.

\subsection{Sampling and quadrature}\label{sec:so3:sampling}

Position is sampled at the $\Npix = 12 N_{\rm side}^2$ \healpix{} pixel centres $x_i$ \citep{gorski2005}, $N_{\rm side}$ being the resolution parameter, and orientation at $\Ngam$ equispaced angles $\gamma_a = 2\pi a/\Ngam$, $a = 0, \dots, \Ngam - 1$; the product is the equivolumetric SO(3) grid of \citet{yershova2010}, and a discrete signal is the set of values $F(x_i,\gamma_a)$. The integral in equation~(\ref{eq:conv}) is replaced by the weighted sum $\sum_i w_i \, (2\pi/\Ngam) \sum_a$ over these points, with weights $w_i$ for the sphere and $2\pi/\Ngam$ for the circle of orientations, while the filter and the group action are kept continuous. This is the discrete-continuous approach of \citet{ocampo2023}. With the filter evaluated exactly at every sample, the only remaining error is that of the quadrature, so equivariance is exact whenever the quadrature rules integrate the discretized integrand exactly (Section~\ref{sec:so3:exact}).

The orientation samples integrate each Fourier mode exactly: for integer $q$ with $|q| < \Ngam$,
\begin{equation}
\frac{1}{\Ngam} \sum_{a=0}^{\Ngam-1} e^{\imagunit q \gamma_a} = \delta_{q0} .
\label{eq:roots}
\end{equation}
For the sphere we require the $w_i$ to integrate exactly every function whose spherical harmonic expansion contains no degree above a band limit $L_w$, i.e. to satisfy the moment conditions $\sum_i w_i \bar{Y}_{\ell m}(x_i) = \sqrt{4\pi} \, \delta_{\ell 0} \delta_{m 0}$ for all $\ell \le L_w$ and $|m| \le \ell$, where $Y_{\ell m}$ are the spherical harmonics. \healpix{} pixels have equal area, so the natural weights $w_i = 4\pi/\Npix$ satisfy these conditions only approximately. We solve for quadrature weights using an initialization of $4\pi/\Npix$ for all pixels \citep{keiner2008} at the band limit $L_w = 2 N_{\rm side}$, which produces small residuals of order $10^{-15}$.

A signal on the group is $2\pi$-periodic with respect to orientation and may be written as a Fourier series,
\begin{equation}
\begin{gathered}
F(x,\gamma) = \sum_{n=-N}^{N} \hat{F}_n(x) \, e^{\imagunit n \gamma} , \\
\hat{F}_n(x) = \frac{1}{\Ngam} \sum_{a=0}^{\Ngam-1} F(x,\gamma_a) \, e^{-\imagunit n \gamma_a} ,
\end{gathered}
\label{eq:components}
\end{equation}
where $\hat{F}_n(x)$ is the orientation component of order $n$ and $N$ the orientation band limit; the second expression is exact, by equation~(\ref{eq:roots}), provided $\Ngam > 2N$. We store signals as their components rather than their samples \citep[compare][]{esteves2020}, and take the range of the filter order $n$ in equation~(\ref{eq:filter}) to be the same $N$. The component $\hat{F}_0$ is the average of $F$ over orientation; it transforms as a scalar field on the sphere, so its average over the sphere is rotation invariant and provides the invariant readout of the network, as explained in Section~\ref{sec:so3:lastlayer}.

\subsection{The discrete layers}\label{sec:so3:layers}

Inserting equations~(\ref{eq:filter}) and (\ref{eq:components}) into the discretized equation~(\ref{eq:conv}), with $u$ and $v$ from equation~(\ref{eq:uv}), the sum over the input orientation samples $\gamma_{a'}$ contains the factor $(2\pi/\Ngam) \sum_{a'} e^{\imagunit (n'-n) \gamma_{a'}}$, where $n'$ is the order of the input component and $n$ the order of the filter in $v$. By equation~(\ref{eq:roots}) this equals $2\pi$ if $n' = n$ and vanishes otherwise, since $|n' - n| \le 2N < \Ngam$, so the orientation integral can be evaluated in closed form. This is how we achieve exact SO(3)-equivariance without onerous computational demands. The output of the convolution, in real-space, is give as
\begin{equation}
H(x_j,\gamma) = \sum_{m=-M}^{M} e^{-\imagunit m \gamma} \; 2\pi \sum_{p=1}^{P} \sum_{n=-N}^{N} c_{pmn} 
\times \!\!\sum_{i:\, \theta_{ij} \le \theta_{\rm c}}\!\! w_i \, R_p(\theta_{ij}) \, e^{\imagunit m \alpha_{j \to i}} \, e^{\imagunit n \alpha_{i \to j}} \, \hat{F}_n(x_i) .
\label{eq:layer_full}
\end{equation}
Comparing with equation~(\ref{eq:components}), $H$ is band limited in orientation and its component of order $k$, with $|k| \le M$, is the coefficient of $e^{\imagunit k \gamma}$, i.e. the terms with $m = -k$. The convolution output, in Fourier-space, is calculated via
\begin{equation}
\hat{H}_k(x_j) = 2\pi \sum_{p=1}^{P} \sum_{n=-N}^{N} c_{p,-k,n} 
\times \!\!\sum_{i:\, \theta_{ij} \le \theta_{\rm c}}\!\! w_i \, R_p(\theta_{ij}) \, e^{-\imagunit k \alpha_{j \to i}} \, e^{\imagunit n \alpha_{i \to j}} \, \hat{F}_n(x_i) .
\label{eq:layer}
\end{equation}
This is the SO(3) to SO(3) layer.  We note that we take $M = N$ so that the output can be passed to the next layer. The lifting layer is the case noted after equation~(\ref{eq:lift}): a spherical input $f$ is the signal whose only non-zero component is $\hat{F}_0 = f$, so equation~(\ref{eq:layer}) retains only the $n = 0$ term and the coefficients $c_{p,-k,0}$. The innermost sum over neighbouring pixels contains no learnable parameters and depends only on the \healpix{} geometry (the separations, bearings, and radial profiles of each pair of neighbouring pixels). These quantities are expensive to hold in memory for every pair but quick to compute, so we regenerate them on the fly at each forward pass rather than store them. Pointwise non-linearities act on the $\Ngam$ orientation samples given by the first expression in equation~(\ref{eq:components}), after which the components are recovered by the second expression.

\subsection{Equivariance of the discrete layers}\label{sec:so3:exact}

The continuous convolution of equation~(\ref{eq:conv}) is equivariant by construction. Here we show that the discrete layer of equation~(\ref{eq:layer}) inherits this, exactly, whenever the two quadrature rules of Section~\ref{sec:so3:sampling} reproduce the integral. There are two conditions. The first relates to the orientation sum, in that the integrand contains frequencies up to $2N$, so by equation~(\ref{eq:roots}) it is integrated exactly provided $\Ngam > 2N$, which we ensure by our choice of $\Ngam$. The second relates to the sum over the sphere. At fixed output pixel $j$, the integrand is the product of the input component, band limited at some $L_f$, and the filter factor, which we take to be band limited at some effective $L_\psi$, so the sum equals the integral provided $L_f + L_\psi \le L_w$. When both these conditions hold, the layer satisfies equation~(\ref{eq:equiv}) exactly.

Since the network stores signals as their orientation components (equation~(\ref{eq:components})), we now show that the components output by a layer transform under rotation in the same way as its input components, so that the output of one layer is a valid input to the next. By equation~(\ref{eq:action}), $(Q \cdot F)(Qx,\gamma) = F(x, \gamma - \chi_Q(x))$, and comparing Fourier coefficients in equation~(\ref{eq:components}),
\begin{equation}
\widehat{(Q \cdot F)}_n(Qx) = e^{-\imagunit n \chi_Q(x)} \, \hat{F}_n(x) .
\label{eq:phases}
\end{equation}
Each component therefore transforms as a spin-weighted field of order $n$ (i.e. it moves with the rotation and picks up a phase set by the transport angle $\chi_Q$). For layers to be stacked, the output components of a layer must transform in the same way.

Under the conditions above the sum over $i$ in equation~(\ref{eq:layer}) equals the corresponding integral, so we can rotate the input and follow how each factor in the summand changes. The separation $\theta_{ij}$ is unchanged by the rotation. The input component $\hat{F}_n(x_i)$ picks up the phase $e^{-\imagunit n \chi_Q(x_i)}$, while the bearing $\alpha_{i \to j}$, being measured from local north at $x_i$, increases by $\chi_Q(x_i)$, so that the factor $e^{\imagunit n \alpha_{i \to j}}$ picks up $e^{\imagunit n \chi_Q(x_i)}$. These two phases cancel because they carry the same order $n$, which is guaranteed by the orthogonality of the orientation modes that collapsed the orientation integral in Section~\ref{sec:so3:layers}. The remaining bearing $\alpha_{j \to i}$ is measured at the output pixel and increases by $\chi_Q(x_j)$, so the factor $e^{-\imagunit k \alpha_{j \to i}}$ picks up $e^{-\imagunit k \chi_Q(x_j)}$. This phase is the same for every term in the sum and factors out, giving
\begin{equation}
\widehat{(Q \cdot H)}_k(Qx_j) = e^{-\imagunit k \chi_Q(x_j)} \, \hat{H}_k(x_j) .
\label{eq:phases_out}
\end{equation}
The output components therefore transform as in equation~(\ref{eq:phases}), with order $k$, so the output of one layer is a valid input to the next.

The requirement that output features transform in the same way as input features is the kernel constraint for rotation-equivariant networks, solved for the plane by \citet{weiler2019} and on curved surfaces by \citet{dehaan2021}; here it is not imposed on the filter but is satisfied automatically by the construction. The sampling conditions above are analogous to those of the SO(3) sampling theorems \citep{kostelec2008, mcewen2015}, with \healpix{} and the weights of Section~\ref{sec:so3:sampling} playing the role of the rule for the polar angle.

These conditions apply to the convolution alone. Pointwise non-linearities, activation functions, broaden the spectrum of the signal in both orientation and position, so that the signal is no longer band limited. This is likely the source of the residual per-cent-level error in Table~\ref{tab:method_comparison}, and it can be mitigated by raising $\Ngam$ and the pixel resolution. We have tested this empirically – decreasing $\Ngam$ increases the error as expected.

\subsection{Invariance for regression/classification tasks}\label{sec:so3:lastlayer}

For tasks like denoising, a network built from these layers is itself approximately equivariant, the residual error coming from the pointwise non-linearities (Section~\ref{sec:so3:exact}). For tasks like regression or classification, where the network outputs a vector (so not matching the input spherical geometry), it does not make sense to consider equivariance of the entire network. Instead, we want a network $\mathcal{F}$ network to be invariant under a transformation $\mathcal{Q}$ for an input $\mathcal{S}$: $\mathcal{F}(\mathcal{Q} \mathcal{S}) = \mathcal{F}(\mathcal{S}) $. To achieve this, the final network architecture's layers, or \textit{head}, must be constructed to preserve SO(3)-invariance. We achieve this through global average pooling of each channel (i.e. averaging all elements in a channel) in advance of the final multi-layer perception (MLP) – this pooling erases any position or rotation information, as desired.

\section{Benchmarking and data}

\begin{table}[t]
\centering
\caption{Comparison of convolution methods for spherical data. Columns 1--2 are described in Section~\ref{app:benchmarks}. Columns 3--5 are validation RMSE for CMB foreground removal and denoising, dense gradient regression and weak lensing parameter estimation respectively; $\langle\mathrm{FOM}\rangle$ is the mean joint posterior figure of merit for $(\Omega_{\mathrm{m}}, S_8)$ inference. We use $\Ngam=8$, achieving per-cent-level $SO(3)$ invariance error at $N_{\mathrm{side}}=256$ – our error can be further reduced to $\mathbf{10^{-4}}$ by introducing a spherical harmonic band limit (we leave the $1.39\%$ value as a simpler direct comparison with the other methods).}
\label{tab:method_comparison}

\setlength{\tabcolsep}{3.5pt}%
\renewcommand{\arraystretch}{1.15}%

\resizebox{0.90\linewidth}{!}{%
\begin{tabular}{@{}lcccccc@{}}
\toprule
Method
& \makecell{Invariance \\ SO(3) err.\ (\%)}
& \makecell{Time\\(ms)}
& \makecell{CMB\\RMSE}
& \makecell{$|\nabla\delta|$\\RMSE}
& \makecell{$\{\Omega_{\mathrm{m}}, S_8\}$\\RMSE}
& \makecell{$\{\Omega_{\mathrm{m}}, S_8\}$\\$\langle\mathrm{FOM}\rangle$} \\
\midrule
Planar
& 107.22
& $\mathbf{0.71{\pm}0.01}$
& $7.490{\times}10^{-1}$
& $4.37{\times}10^{-2}$
& $4.862{\times}10^{-2}$
& 561 \\
\addlinespace
DISCO
& $2.36$
& $1.93{\pm}0.02$
& $4.212{\times}10^{-1}$
& $4.08{\times}10^{-2}$
& $2.623{\times}10^{-2}$
& 1724 \\
\addlinespace
DeepSphere
& $71.56$
& $54.86{\pm}0.02$
& $\mathbf{1.755{\times}10^{-1}}$
& $9.80{\times}10^{-3}$
& $2.291{\times}10^{-2}$
& 3467 \\
\addlinespace
\textbf{This work}
& $\mathbf{1.39}$
& $27.91{\pm}1.13$
& $2.261{\times}10^{-1}$
& $\mathbf{4.43{\times}10^{-3}}$
& $\mathbf{1.651{\times}10^{-2}}$
& \textbf{5026} \\
\bottomrule
\end{tabular}%
}
\end{table}

\subsection{Simulation-based inference}
 \label{sec:sbi}

For a learned neural network compression of a weak lensing map the likelihood $p(d \mid \theta)$ has no closed form, but a simulator can draw $d \sim p(d \mid \theta)$. Simulation-based inference (SBI) learns the posterior from such simulated pairs \citep{Cranmer30055}. Density estimation in $\mathcal{O}(\Npix)$ dimensions is not feasible, so a network first compresses each map to a low-dimensional summary $\mathbf{t} = F(d)$, whose dimension is usually of the same order as the number of model parameters. This compressor is where the convolution of Section~\ref{sec:so3} enters. We use neural posterior estimation \citep{papamakarios2018fastepsilonfreeinferencesimulation}: a masked autoregressive flow $q_\varphi(\theta \mid \mathbf{t})$ \citep[MAF;][]{papa_mafs2017} is fitted to the simulated pairs $(\theta_i, \mathbf{t}_i)$ by minimizing $-\sum_i \ln q_\varphi(\theta_i \mid \mathbf{t}_i)$. The fitted flow is amortized, yielding a posterior $q_\varphi(\theta \mid \mathbf{t}_{\mathcal{O}})$ for the summary $\mathbf{t}_{\mathcal{O}} = F(d_{\mathcal{O}})$ of any new observation $d_{\mathcal{O}}$ with no further simulation and no MCMC.

\subsection{Dense gradient-magnitude regression task}
 \label{sec:grf_task}

Each input is a Gaussian random field $\delta$ on the unit sphere $\mathbb{S}^2$, generated in harmonic space by drawing its spherical harmonic coefficients $a_{\ell m}$ as independent zero-mean Gaussian variables with variance $C_\ell$, where $\ell$ is the multipole and $C_\ell$ the angular power spectrum. We use the smooth power-law spectrum
\begin{equation}
\label{eq:grf_cl}
C_\ell = A\,(\ell+10^{-3})^{-\alpha}\exp\!\big[-(\ell/\ell_{\mathrm{bl}})^2\big] , \qquad C_0 = 0 ,
\end{equation}
where $\alpha = 1.8$ is the spectral slope, the Gaussian factor suppresses power above the multipole $\ell_{\mathrm{bl}} = 2N_{\rm side}$, and the small offset $10^{-3}$ keeps the power law finite at $\ell = 0$ (the monopole is in any case set to zero, so every map has zero mean). The amplitude $A$ is fixed in closed form by requiring $\mathrm{Var}[\delta] = \sum_\ell (2\ell+1) C_\ell / 4\pi = 1$, so that every map has unit variance. The maps are band limited at $\ell_{\max} = 2N_{\rm side}$ (i.e. $a_{\ell m} = 0$ for $\ell > \ell_{\max}$), matching the band limit $L_w$ of our quadrature weights (Section~\ref{sec:so3:sampling}).

The target is the smoothed gradient magnitude
\begin{equation}
\label{eq:grf_grad}
|\nabla\delta| = \sqrt{(\partial_\theta\delta)^2 + \frac{1}{\sin^2\theta}(\partial_\phi\delta)^2} ,
\end{equation}
where $\theta$ and $\phi$ are the colatitude and longitude of Section~\ref{sec:so3:rot}. The partial derivatives are computed analytically from the $a_{\ell m}$ rather than by finite differencing of the pixelized map, avoiding any discretization error in the target.

\subsection{Cosmic Microwave Background denoising}
 \label{sec:cmb_denoise}
We evaluate on the CMB-ML benchmark \citep{amato2025cmbml}, which provides simulated Planck-like sky maps at nine frequencies ($N_{\mathrm{side}}=256$): a lensed CMB realization combined with Galactic and extragalactic foreground emission and correlated, frequency-dependent instrumental noise. The task is to recover the underlying CMB temperature map from these nine contaminated observations (Fig.~\ref{fig:cmb}), a spherical analogue of denoising under structured, spatially correlated corruption rather than i.i.d.\ pixel noise.

\begin{figure}[t]
  \centering
  \includegraphics[width=0.8\linewidth]{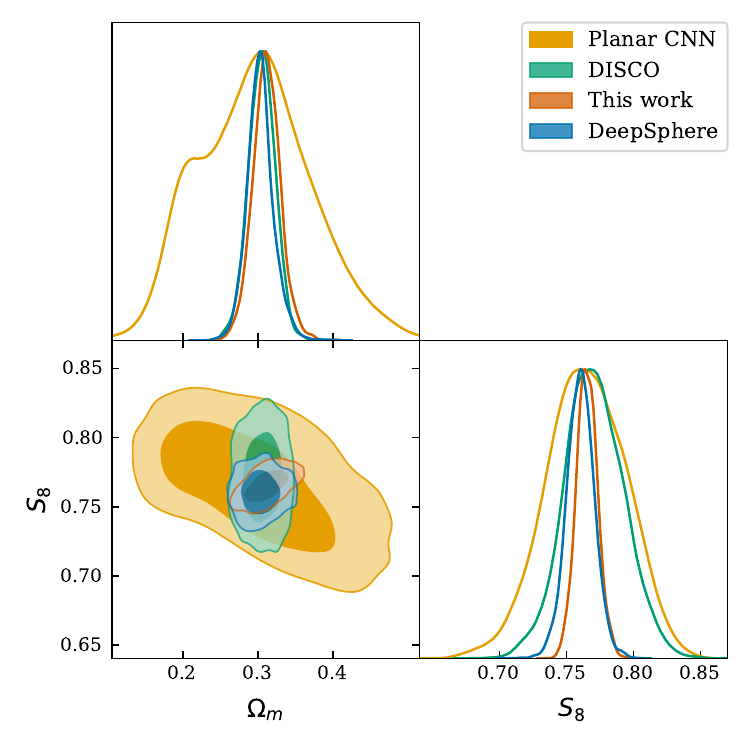}
  \caption{
    Cosmological parameter $\Omega_{\mathrm{m}}$--$S_8$
    posteriors from simulation-based inference with lensing data,
    comparing a CNN from this work with a planar CNN, DISCO and
    DeepSphere compressors (i.e.\ embeddings) using the same number
    of expensive training simulations. The poor results from models not optimized for spherical data (especially the Planar CNN) given a fixed simulation budget forces
    a user to rely on imperfect simulators to increase the simulation
    count -- our method avoids this by
    including a spherical symmetry as inductive bias.
  }
  \label{fig:om_s8_sbi}
\end{figure}

\subsection{Cosmological parameters from weak lensing maps via simulation-based inference}
 \label{sec:om_s8_task}

We use the Gower Street suite of full-sky simulations \citep{jeffrey2024dark}, from which we build a weak lensing convergence map at $N_{\mathrm{side}}=256$ for each realization (Fig.~\ref{fig:kappa}), with a single source plane and a fixed source redshift; the construction is described in Section~\ref{app:om_s8_data}. Unlike the pixel-wise targets of the previous two tasks, the regression target here is a pair of numbers: the matter density $\Omega_{\rm m}$ and the clustering amplitude $S_8$, which is the combination of parameters that weak lensing constrains most tightly.

The summary given to the flow is the two-dimensional point estimate $\mathbf{t} = (\hat\Omega_{\rm m}, \hat S_8)$ returned by the regression network under its mean squared error loss, so the density estimator is conditioned on two numbers per map rather than on the $\kappa$ map itself. We fit a single MAF, with four networks in the ensemble, using two autoregressive transforms and eight hidden features per transform, implemented with the \texttt{sbi} package \citep{tejerocantero2020sbi}. It is trained on the parameter-summary pairs from the simulations, with the 78 maps used for the NDE validation loss. This choice comes from the limited number of simulations; as no overfitting is observed in any architecture, this does not induce a bias in our results. The joint figure of merit (FOM) in Table~\ref{tab:method_comparison} is computed from the resulting average FOM of the 78 held out validation posteriors.

\subsection{Weak lensing convergence maps}
\label{app:om_s8_data}

This section gives the details of the data used for the cosmological parameter regression task of Section~\ref{sec:om_s8_task}. The Gower Street suite is a set of 791 full-sky $N$-body simulations of a seven-parameter $\nu w$CDM cosmology, run with \textsc{PKDGRAV3} \citep{potter2017pkdgrav3} and described in \citet{jeffrey2024dark}. Each simulation provides pixelized matter overdensity shells $\delta(\hat n,\chi)$ along the observer's lightcone, where $\hat n$ is a direction on the sky and $\chi$ is comoving distance. From these shells we build a Born-approximated convergence map for a single source plane fixed at $z=0.8$,
\begin{equation}
\label{eq:convergence_task}
\kappa(\hat n) = \frac{3\Omega_{\rm m}H_0^2}{2c^2}\int_0^{\chi_s}\frac{\chi(\chi_s-\chi)}{\chi_s}\,\frac{\delta(\hat n,\chi)}{a(\chi)}\,\mathrm{d}\chi,
\end{equation}
with $\chi_s$ the comoving distance to the source plane, $a$ the scale factor, $H_0$ the Hubble constant and $c$ the speed of light. Every map is then downgraded to $N_{\mathrm{side}}=256$.

The two regression targets are the matter density $\Omega_{\rm m}$ and the clustering amplitude $S_8\equiv\sigma_8(\Omega_{\rm m}/0.3)^{0.5}$. The five remaining parameters varied across the suite, namely $w$, $\Omega_{\rm b}h^2$, $m_\nu$, $h$ and $n_s$, are not predicted by the network and are therefore marginalized over implicitly.

\subsection{Networks}
 \label{sec:networks}

We compare the convolution of Section~\ref{sec:so3} against three alternatives. The architectures and network configurations for the weak lensing parameter estimation task are given in Section~\ref{app:architectures}. The network of this work uses band limits $P=4$, $M=N=1$, and $\Ngam=8$ orientation samples. The radial profiles $R_{p,|m|}$ are Zernike radial polynomials of order $|m|+2p$ in the angular separation $\theta_{ij}$, multiplied by a window function that falls smoothly to zero at the cutoff $\theta_{\rm c} = 1.6\times10^{-2}\,\mathrm{rad}$, and all computations are performed in single precision.

\subsubsect{DeepSphere} The \healpix{} grid is treated as a graph, with pixel centres as vertices and edge weights given by a Gaussian kernel of the Euclidean distance between the $k$ nearest neighbours, and the convolution is a polynomial in the resulting graph Laplacian \citep{defferrard2016, perraudin2019, defferrard2020}. It costs $\mathcal{O}(\Npix)$ and pools along the \healpix{} nested pixelization hierarchy, its filters are isotropic and the graph Laplacian produces asymptotic equivariance only as $\Npix \to \infty$, so equivariance is approximate at finite resolution.

\subsubsect{DISCO} The signal and the quadrature are discretized on the sphere, while the filter and the rotation acting on it stay continuous \citep{ocampo2023}, as in equation~(\ref{eq:layer}), and we use the DISCO implementation of \citet{bonev2023}, where we port the convolutions to a \healpix{} sampling scheme. Rotations are restricted to the quotient space $\mathrm{SO}(3)/\mathrm{SO}(2)$, with a non-directional axisymmetric filter, so that it is $\mathrm{SO}(3)$ equivariant up to the accuracy of its quadrature. We use only this axisymmetric $\mathrm{SO}(3)/\mathrm{SO}(2)$ form, as it is the only one available in the public DISCO code;\footnote{\url{https://github.com/NVIDIA/torch-harmonics}} full $\mathrm{SO}(3)$ DISCO convolutions with directional filters are, in any case, considerably more expensive.

\subsubsect{Planar CNN} The sphere is unrolled onto a single equirectangular projection whose rows are the $4N_{\rm side}-1$ iso-latitude \healpix{} rings and whose columns are the $4N_{\rm side}$ pixels of the equator, so that near-pole rings holding as few as four distinct pixels are stretched across every column. An ordinary two-dimensional convolution stack is applied to this raster, periodic in longitude, zero padded in latitude and read out through a solid-angle weighted pool. It encodes no rotational symmetry and heavily distorts the true spherical geometry. Every architecture is trained either to convergence or with early stopping on the validation loss to prevent overfitting. We use an 80/20 training-validation split for all datasets, apart from the weak lensing maps of Section~\ref{sec:om_s8_task}, where the split is 90/10.

\subsection{Network architectures}
\label{app:architectures}
This section gives the parameter budget of the four networks compared on the weak lensing task of Section~\ref{sec:om_s8_task}, as totals in Table~\ref{tab:param_totals} and layer by layer in Tables~\ref{tab:params_planar}--\ref{tab:params_so3}. All four use 16 hidden channels at a single resolution, $N_{\mathrm{side}}=256$, and read out the two parameters $(\Omega_{\rm m}, S_8)$ through a linear head. Operations carrying no learnable parameters are omitted: activations, pooling, normalization by quadrature, and the fixed geometric quantities of each convolution, which are held as non-persistent buffers (the resampling indices and solid-angle pooling weights of the planar raster, the graph Laplacian of DeepSphere, and the separations, bearings and radial profiles of equation~(\ref{eq:layer})).
\begin{table}[htbp]
\centering
\caption{Total learnable parameters. The convolution of Section~\ref{sec:so3} has more learnable parameters than the other three networks (a factor of 11 more than DISCO and about 2.5 more than the planar CNN and DeepSphere), since its filter coefficients are complex and indexed by radial profile and by both bearing orders.}
\label{tab:param_totals}
\begin{tabular}{@{}lr@{}}
\toprule
Network & Total params \\
\midrule
This work    & 27{,}938 \\
Planar CNN   & 11{,}794 \\
DeepSphere   & 10{,}770 \\
DISCO        & \phantom{0}2{,}450 \\
\bottomrule
\end{tabular}
\end{table}
\begin{table}[htbp]
\centering
\caption{Learnable parameters of the planar CNN. Each convolution is padded manually, periodically in longitude and with zeros in colatitude.}
\label{tab:params_planar}
\setlength{\tabcolsep}{5pt}%
\begin{tabular}{@{}llccr@{}}
\toprule
Layer & Type & Weight shape & Bias shape & \# Params \\
\midrule
\texttt{blocks.0} & Conv2d ($3{\times}3$) & $(16,1,3,3)$  & $(16)$ & 160 \\
\texttt{blocks.1} & Conv2d ($3{\times}3$) & $(16,16,3,3)$ & $(16)$ & 2{,}320 \\
\texttt{blocks.2} & Conv2d ($3{\times}3$) & $(16,16,3,3)$ & $(16)$ & 2{,}320 \\
\texttt{blocks.3} & Conv2d ($3{\times}3$) & $(16,16,3,3)$ & $(16)$ & 2{,}320 \\
\texttt{blocks.4} & Conv2d ($3{\times}3$) & $(16,16,3,3)$ & $(16)$ & 2{,}320 \\
\texttt{blocks.5} & Conv2d ($3{\times}3$) & $(16,16,3,3)$ & $(16)$ & 2{,}320 \\
\texttt{head}     & Linear                & $(2,16)$      & $(2)$  & 34 \\
\midrule
\textbf{Total} & & & & \textbf{11{,}794} \\
\bottomrule
\end{tabular}
\end{table}
\begin{table}[htbp]
\centering
\caption{Learnable parameters of DISCO, with an axisymmetric filter of three radial taps.}
\label{tab:params_disco}
\setlength{\tabcolsep}{5pt}%
\begin{tabular}{@{}llccr@{}}
\toprule
Layer & Type & Weight shape & Bias shape & \# Params \\
\midrule
\texttt{lift}     & Discrete-continuous conv & $(16,1,3)$  & $(16)$ & 64 \\
\texttt{blocks.0} & Discrete-continuous conv & $(16,16,3)$ & $(16)$ & 784 \\
\texttt{blocks.1} & Discrete-continuous conv & $(16,16,3)$ & $(16)$ & 784 \\
\texttt{blocks.2} & Discrete-continuous conv & $(16,16,3)$ & $(16)$ & 784 \\
\texttt{head}     & Linear                   & $(2,16)$    & $(2)$  & 34 \\
\midrule
\textbf{Total} & & & & \textbf{2{,}450} \\
\bottomrule
\end{tabular}
\end{table}
\begin{table}[htbp]
\centering
\caption{Learnable parameters of DeepSphere, with Chebyshev order $K=4$ and a $1{\to}16{\to}32{\to}64$ channel schedule, pooling between stages. Layer names are relative to the encoder, whose stages wrap the convolution and its normalization together.}
\label{tab:params_deepsphere}
\setlength{\tabcolsep}{5pt}%
\begin{tabular}{@{}llccr@{}}
\toprule
Layer & Type & Weight shape & Bias shape & \# Params \\
\midrule
\texttt{blocks.0} & ChebConv ($K{=}4$) & $(4,1,16)$  & $(16)$ & 80 \\
\texttt{blocks.0} & BatchNorm1d        & $(16)$      & $(16)$ & 32 \\
\texttt{blocks.1} & ChebConv ($K{=}4$) & $(4,16,32)$ & $(32)$ & 2{,}080 \\
\texttt{blocks.1} & BatchNorm1d        & $(32)$      & $(32)$ & 64 \\
\texttt{blocks.2} & ChebConv ($K{=}4$) & $(4,32,64)$ & $(64)$ & 8{,}256 \\
\texttt{blocks.2} & BatchNorm1d        & $(64)$      & $(64)$ & 128 \\
\texttt{head}     & Linear             & $(2,64)$    & $(2)$  & 130 \\
\midrule
\textbf{Total} & & & & \textbf{10{,}770} \\
\bottomrule
\end{tabular}
\end{table}
\begin{table}[htbp]
\centering
\caption{Learnable parameters of the network of this work. The band limits of equation~(\ref{eq:filter}) give four radial profiles, up to two orders and three taps per channel pair; the coefficients $c_{pmn}$ are complex, so real and imaginary parts are stored separately and the imaginary part of the $m=0$ order is absent.}
\label{tab:params_so3}
\setlength{\tabcolsep}{5pt}%
\begin{tabular}{@{}llccr@{}}
\toprule
Layer & Type & Weight shapes (re, im) & Bias shape & \# Params \\
\midrule
\texttt{lift}     & Lift $S^2 \to \mathrm{SO}(3)$ & $(16,1,4,2)$, $(16,1,4,1)$       & $(16)$ & 208 \\
\texttt{blocks.0} & $\mathrm{SO}(3)$ conv         & $(16,16,4,2,3)$, $(16,16,4,1,3)$ & $(16)$ & 9{,}232 \\
\texttt{blocks.1} & $\mathrm{SO}(3)$ conv         & $(16,16,4,2,3)$, $(16,16,4,1,3)$ & $(16)$ & 9{,}232 \\
\texttt{blocks.2} & $\mathrm{SO}(3)$ conv         & $(16,16,4,2,3)$, $(16,16,4,1,3)$ & $(16)$ & 9{,}232 \\
\texttt{head}     & Linear                        & $(2,16)$                         & $(2)$  & 34 \\
\midrule
\textbf{Total} & & & & \textbf{27{,}938} \\
\bottomrule
\end{tabular}
\end{table}

\subsection{Invariance and time tests}
\label{app:benchmarks}

The invariance error of Table~\ref{tab:method_comparison} is measured end-to-end on the trained networks, which are comprised of our equivariant convolutional layers, of the weak lensing task. Each validation map is rotated on the sphere by a random Euler rotation, and both the original and rotated maps are normalized by the training-set statistics and passed through the same frozen network. The error is the percentage deviation of the predicted summaries under rotation, and Table~\ref{tab:method_comparison} quotes the mean over rotations and validation maps. The rotations are drawn once from a fixed seed and reused across every architecture.

The time column of Table~\ref{tab:method_comparison} times the first layer of each network, which maps the raw \healpix{} scalar map into the working representation of the given architecture, rather than the full forward pass, since the networks differ in depth and a whole-model timing would largely measure those unrelated choices. All four architectures are timed sequentially in a single process on an otherwise idle NVIDIA A100 GPU. A single $N_{\mathrm{side}} = 256$ map is passed at batch size one; warm-up iterations are discarded to settle kernel selection, then 50 forward passes and 50 forward--backward passes are timed individually, so the quoted numbers are per-call latencies rather than throughput. Table~\ref{tab:method_comparison} reports the mean and standard deviation of the forward pass in single precision.

\subsection{Results}
 \label{sec:results}

Referring to Table~\ref{tab:method_comparison}, as expected the planar architecture has poor $\mathrm{SO}(3)$ equivariance. DeepSphere also exhibits large equivariance error, due to the relatively poor resolution of the input maps. Our implementation, along with DISCO, offers near per-cent-level equivariance for the entire network. Under random rotations of the map, the inferred summaries shift on average only around a couple of per cent. The learned summaries are therefore built from features that are $\mathrm{SO}(3)$ equivariant to per-cent-level accuracy.

The usefulness of this is evidenced by the weak lensing parameter estimation task, where our architecture both provides the best validation loss and the largest average FOM, thereby resulting in the most precise constraints of all the methods. Fig.~\ref{fig:om_s8_sbi} shows the posteriors of each method for a map at the prior mean of $\Omega_{\rm m}$ and $S_8$ from the Gower Street simulations.

We attribute this success not only to the exact equivariance of its convolutional layers but also to the directionality of its filters: the non-zero bearing orders $m$ and $n$ of equation~(\ref{eq:filter}) carry oriented, spin-weighted features that an axisymmetric filter cannot represent. We note, however, that the $\kappa$ maps are of relatively low resolution ($N_{\mathrm{side}}=256$), so the amount of non-linear (non-Gaussian) information available to be extracted beyond the power spectrum is limited. The planar CNN collapses to the trivial solution of predicting the ensemble mean of the dataset, so its posteriors in $\Omega_{\rm m}$ are prior driven. The planar CNN cannot predict $\Omega_{\rm m}$ at the resolution of the input maps.

In the remaining tasks of CMB denoising and dense gradient regression on Gaussian random fields, the spherical convolution methods all perform significantly better than the naive planar method. Our method performs particularly well in the Gaussian random field dense gradient regression task. DeepSphere performs best in the CMB denoising task. We note that this task is not strictly $\mathrm{SO}(3)$ equivariant, as the instrumental noise has a preferred orientation on the sky, so an exact symmetry inductive bias is not necessarily optimal here.

\section{Discussion}
 \label{sec:discussion}

The method has limitations. It is slower than the planar and DISCO baselines (Table~\ref{tab:method_comparison}). Its equivariance is exact only under the assumptions of Section~\ref{sec:so3}, which real observations satisfy only approximately.

In this paper we have presented a SO(3)-equivariant convolutional neural network and introduced four distinct contributions.

First, in Section~\ref{sec:so3} we have introduced a convolution for data on the sphere, defined directly on the \healpix{} grid, that is exactly equivariant under $\mathrm{SO}(3)$, the group of three-dimensional rotations: rotating the input map rotates the output of the convolution in the same way. The symmetry is built into the structure of the network itself, through steerable filters, parallel transport of local reference frames and spin-weighted features, rather than being learned from training data.

Second, this exactness does not come at prohibitive cost: each filter acts only over a small neighbourhood of each pixel, no transform of the full sky is ever required, and in the tests of Section~\ref{app:benchmarks} the predictions of the trained networks shift by only around one per cent when the input maps are rotated, the smallest deviation of all the methods compared. 

Third, on tasks where the true answer is known, the method recovers the gradient magnitude of Gaussian random fields (Section~\ref{sec:grf_task}) more accurately than the planar, graph-based, and axisymmetric DISCO alternatives, and removes noise and foreground contamination from cosmic microwave background maps (Section~\ref{sec:cmb_denoise}) with accuracy comparable to other methods in the literature.

Fourth, in Section~\ref{sec:om_s8_task} we have shown what the symmetry is worth for inference: given the same number of training simulations, our network yields the tightest and most accurate constraints on the cosmological parameters $\Omega_{\rm m}$ and $S_8$, the matter density and the amplitude of matter clustering from cosmology data. A given measurement precision therefore requires fewer training simulations, and a fixed budget can be spent making each simulation more faithful to reality rather than simply making more of them, reducing the reliance on imperfect models in scientific analyses.
We have introduced a scalable, exactly $\mathrm{SO}(3)$ equivariant convolution on the \healpix{} grid and demonstrated it in SBI from spherical cosmological data. The symmetry is supplied by the architecture rather than learned from data, so a given precision requires fewer training simulations. 

% ============================================================
% End matter
% ============================================================

\ack{ The authors thank Ofer Lahav for useful discussions.
JW is supported by the STFC UCL Centre for Doctoral Training in Data Intensive Science (grant ST/W00674X/1) and by departmental and industry contributions. NJ acknowledges support from the King’s AI+ Academic Fellowship at King’s College London. NJ also acknowledges support from the ERC-selected UKRI Frontier Research Grant EP/Y03015X/1.
The authors declare no competing interests.
}

\data{
The data and code will be made publicly available upon acceptance.
}

% ============================================================
% References
% ============================================================

% The "References" heading is generated by the bibliography itself.
\bibliographystyle{iopart-harvard}
\bibliography{references}

@inproceedings{amato2025cmbml,
  author    = {Amato, James and Xie, Yunan and Medina-Varela, Leonel and Aljerwi, Ammar and McCutcheon, Adam and Rippentrop, T. Seth and Gonzalez, Kristian and Delabrouille, Jacques and Ishak, Mustapha and Ruozzi, Nicholas},
  title     = {{CMB-ML}: A Cosmic Microwave Background Dataset for the Oldest Possible Computer Vision Task},
  booktitle = {Proceedings of the IEEE/CVF International Conference on Computer Vision (ICCV)},
  year      = {2025},
  month     = {October},
  pages     = {9418--9430}
}

@article{gorski2005,
  author  = {G{\'o}rski, K. M. and Hivon, E. and Banday, A. J. and Wandelt, B. D. and Hansen, F. K. and Reinecke, M. and Bartelmann, M.},
  title   = {{HEALPix}: A Framework for High-Resolution Discretization and Fast Analysis of Data Distributed on the Sphere},
  journal = {The Astrophysical Journal},
  volume  = {622},
  pages   = {759--771},
  year    = {2005},
  doi     = {10.1086/427976}
}

@article{yershova2010,
  author  = {Yershova, A. and Jain, S. and LaValle, S. M. and Mitchell, J. C.},
  title   = {Generating uniform incremental grids on {SO(3)} using the Hopf fibration},
  journal = {International Journal of Robotics Research},
  year    = {2010},
  volume  = {29},
  number  = {7},
  pages   = {801--812},
  doi     = {10.1177/0278364909352700}
}

@article{keiner2008,
  author  = {Keiner, J. and Potts, D.},
  title   = {Fast evaluation of quadrature formulae on the sphere},
  journal = {Mathematics of Computation},
  year    = {2008},
  volume  = {77},
  number  = {261},
  pages   = {397--419},
  doi     = {10.1090/S0025-5718-07-02029-7}
}

@article{freeman1991,
  author  = {Freeman, W. T. and Adelson, E. H.},
  title   = {The design and use of steerable filters},
  journal = {IEEE Transactions on Pattern Analysis and Machine Intelligence},
  year    = {1991},
  volume  = {13},
  number  = {9},
  pages   = {891--906},
  doi     = {10.1109/34.93808}
}

@inproceedings{worrall2017,
  author    = {Worrall, D. E. and Garbin, S. J. and Turmukhambetov, D. and Brostow, G. J.},
  title     = {Harmonic networks: deep translation and rotation equivariance},
  booktitle = {IEEE Conference on Computer Vision and Pattern Recognition (CVPR)},
  year      = {2017},
  pages     = {5028--5037},
  doi       = {10.1109/CVPR.2017.758}
}

@inproceedings{cohen2016,
  author    = {Cohen, T. S. and Welling, M.},
  title     = {Group equivariant convolutional networks},
  booktitle = {International Conference on Machine Learning (ICML)},
  year      = {2016},
  pages     = {2990--2999}
}

@inproceedings{cohen2018,
  author    = {Cohen, Taco S. and Geiger, Mario and K{\"o}hler, Jonas and Welling, Max},
  title     = {Spherical {CNNs}},
  booktitle = {International Conference on Learning Representations},
  year      = {2018}
}

@inproceedings{cohen2019,
  author    = {Cohen, T. S. and Weiler, M. and Kicanaoglu, B. and Welling, M.},
  title     = {Gauge equivariant convolutional networks and the icosahedral {CNN}},
  booktitle = {International Conference on Machine Learning (ICML)},
  year      = {2019},
  pages     = {1321--1330}
}

@inproceedings{esteves2018,
  author    = {Esteves, Carlos and Allen-Blanchette, Christine and Makadia, Ameesh and Daniilidis, Kostas},
  title     = {Learning {SO(3)} Equivariant Representations with Spherical {CNNs}},
  booktitle = {Proceedings of the European Conference on Computer Vision (ECCV)},
  pages     = {52--68},
  year      = {2018}
}

@inproceedings{esteves2020,
  author    = {Esteves, Carlos and Makadia, Ameesh and Daniilidis, Kostas},
  title     = {Spin-Weighted Spherical {CNNs}},
  booktitle = {Advances in Neural Information Processing Systems},
  volume    = {33},
  pages     = {8614--8625},
  year      = {2020}
}

@inproceedings{esteves2023,
  author    = {Esteves, Carlos and Slotine, Jean-Jacques and Makadia, Ameesh},
  title     = {Scaling Spherical {CNNs}},
  booktitle = {Proceedings of the 40th International Conference on Machine Learning},
  series    = {Proceedings of Machine Learning Research},
  volume    = {202},
  pages     = {9396--9411},
  year      = {2023}
}

@inproceedings{cobb2021,
  author    = {Cobb, Oliver J. and Wallis, Christopher G. R. and Mavor-Parker, Augustine N. and Marignier, Augustin and Price, Matthew A. and d'Avezac, Mayeul and McEwen, Jason D.},
  title     = {Efficient Generalized Spherical {CNNs}},
  booktitle = {International Conference on Learning Representations},
  year      = {2021}
}

@inproceedings{ocampo2023,
  author    = {Ocampo, Jeremy and Price, Matthew A. and McEwen, Jason D.},
  title     = {Scalable and equivariant spherical {CNNs} by discrete-continuous ({DISCO}) convolutions},
  booktitle = {International Conference on Learning Representations},
  year      = {2023}
}

@inproceedings{weiler2019,
  author    = {Weiler, M. and Cesa, G.},
  title     = {General {$E(2)$}-equivariant steerable {CNNs}},
  booktitle = {Advances in Neural Information Processing Systems (NeurIPS)},
  year      = {2019},
  volume    = {32}
}

@article{kostelec2008,
  author  = {Kostelec, P. J. and Rockmore, D. N.},
  title   = {{FFTs} on the rotation group},
  journal = {Journal of Fourier Analysis and Applications},
  year    = {2008},
  volume  = {14},
  number  = {2},
  pages   = {145--179},
  doi     = {10.1007/s00041-008-9013-5}
}

@article{mcewen2015,
  author  = {McEwen, J. D. and B{\"u}ttner, M. and Leistedt, B. and Peiris, H. V. and Wiaux, Y.},
  title   = {A novel sampling theorem on the rotation group},
  journal = {IEEE Signal Processing Letters},
  year    = {2015},
  volume  = {22},
  number  = {12},
  pages   = {2425--2429},
  doi     = {10.1109/LSP.2015.2490676}
}

@article{perraudin2019,
  author  = {Perraudin, Nathana{\"e}l and Defferrard, Micha{\"e}l and Kacprzak, Tomasz and Sgier, Rapha{\"e}l},
  title   = {{DeepSphere}: Efficient spherical convolutional neural network with {HEALPix} sampling for cosmological applications},
  journal = {Astronomy and Computing},
  volume  = {27},
  pages   = {130--146},
  year    = {2019},
  doi     = {10.1016/j.ascom.2019.03.004}
}

@inproceedings{defferrard2020,
  author    = {Defferrard, Micha{\"e}l and Milani, Martino and Gusset, Fr{\'e}d{\'e}rick and Perraudin, Nathana{\"e}l},
  title     = {{DeepSphere}: a graph-based spherical {CNN}},
  booktitle = {International Conference on Learning Representations},
  year      = {2020}
}

@article{jeffrey2024dark,
  author  = {Jeffrey, N. and Whiteway, L. and Gatti, M. and Williamson, J. and Alsing, J. and Porredon, A. and others},
  title   = {Dark Energy Survey Year 3 results: likelihood-free, simulation-based {$w$CDM} inference with neural compression of weak-lensing map statistics},
  journal = {Monthly Notices of the Royal Astronomical Society},
  volume  = {536},
  pages   = {1303},
  year    = {2025},
  doi     = {10.1093/mnras/stae2629},
  eprint  = {2403.02314}
}

@article{potter2017pkdgrav3,
  title         = {{PKDGRAV3: beyond trillion particle cosmological simulations for the next era of galaxy surveys}},
  author        = {Potter, Douglas and Stadel, Joachim and Teyssier, Romain},
  year          = {2017},
  month         = may,
  journal       = {Computational Astrophysics and Cosmology},
  volume        = {4},
  number        = {1},
  pages         = {2},
  doi           = {10.1186/s40668-017-0021-1},
  archiveprefix = {arXiv},
  eprint        = {1609.08621}
}

@article{planck2020,
  author  = {{Planck Collaboration} and Aghanim, N. and Akrami, Y. and others},
  title   = {Planck 2018 results. {VI}. Cosmological parameters},
  journal = {Astronomy \& Astrophysics},
  volume  = {641},
  pages   = {A6},
  year    = {2020},
  doi     = {10.1051/0004-6361/201833910}
}

@article{williamson2026,
  author  = {Williamson, J. and Makinen, T. L. and others},
  title   = {Dark Energy Survey Year 3 results: optimized {$w$CDM} simulation-based inference with weak lensing map-level hybrid statistics},
  journal = {arXiv e-prints},
  pages   = {arXiv:2606.11309},
  year    = {2026},
  eprint  = {2606.11309}
}

@article{thomsen2025,
  author  = {Thomsen, A. and others},
  title   = {Dark Energy Survey Year 3 results: Simulation-based {$w$CDM} inference from weak lensing and galaxy clustering maps with deep learning: Analysis design},
  journal = {Physical Review D},
  year    = {2026},
  doi     = {10.1103/3sj1-1l9f},
  eprint  = {2511.04681}
}

@article{wolz2023,
  author  = {Wolz, K. and Krachmalnicoff, N. and Pagano, L.},
  title   = {Inference of the optical depth to reionization {$\tau$} from {Planck} {CMB} maps with convolutional neural networks},
  journal = {Astronomy \& Astrophysics},
  volume  = {676},
  pages   = {A30},
  year    = {2023},
  doi     = {10.1051/0004-6361/202345982},
  eprint  = {2301.09634}
}

@article{hu2025smgno,
  author  = {Hu, Yifan and Yin, Fukang and Zhang, Weimin and Ren, Kaijun and Song, Junqiang and Deng, Kefeng and Zhang, Di},
  title   = {Spherical multigrid neural operator for improving autoregressive global weather forecasting},
  journal = {Scientific Reports},
  volume  = {15},
  pages   = {11522},
  year    = {2025},
  doi     = {10.1038/s41598-025-96208-y}
}

@inproceedings{bonev2023,
  author    = {Bonev, Boris and Kurth, Thorsten and Hundt, Christian and Pathak, Jaideep and Baust, Maximilian and Kashinath, Karthik and Anandkumar, Anima},
  title     = {Spherical {F}ourier Neural Operators: Learning Stable Dynamics on the Sphere},
  booktitle = {Proceedings of the 40th International Conference on Machine Learning},
  series    = {Proceedings of Machine Learning Research},
  volume    = {202},
  pages     = {2806--2823},
  year      = {2023}
}

@inproceedings{kondor2018,
  author    = {Kondor, Risi and Lin, Zhen and Trivedi, Shubhendu},
  title     = {Clebsch--Gordan Nets: a Fully {F}ourier Space Spherical Convolutional Neural Network},
  booktitle = {Advances in Neural Information Processing Systems},
  volume    = {31},
  year      = {2018}
}

@inproceedings{dehaan2021,
  author    = {de Haan, P. and Weiler, M. and Cohen, T. and Welling, M.},
  title     = {Gauge equivariant mesh {CNNs}: anisotropic convolutions on geometric graphs},
  booktitle = {International Conference on Learning Representations (ICLR)},
  year      = {2021}
}

@article{Cranmer30055,
  author  = {Cranmer, Kyle and Brehmer, Johann and Louppe, Gilles},
  title   = {The frontier of simulation-based inference},
  journal = {Proceedings of the National Academy of Sciences},
  volume  = {117},
  number  = {48},
  pages   = {30055--30062},
  year    = {2020},
  doi     = {10.1073/pnas.1912789117}
}

@inproceedings{papamakarios2018fastepsilonfreeinferencesimulation,
  author    = {Papamakarios, George and Murray, Iain},
  title     = {Fast $\epsilon$-free Inference of Simulation Models with Bayesian Conditional Density Estimation},
  booktitle = {Advances in Neural Information Processing Systems},
  volume    = {29},
  year      = {2016},
  eprint    = {1605.06376},
  archivePrefix = {arXiv},
  primaryClass  = {stat.ML}
}

@inproceedings{papa_mafs2017,
  author    = {Papamakarios, George and Pavlakou, Theo and Murray, Iain},
  title     = {Masked Autoregressive Flow for Density Estimation},
  booktitle = {Advances in Neural Information Processing Systems},
  volume    = {30},
  year      = {2017},
  eprint    = {1705.07057},
  archivePrefix = {arXiv},
  primaryClass  = {stat.ML}
}

@article{tejerocantero2020sbi,
  author  = {Tejero-Cantero, Alvaro and Boelts, Jan and Deistler, Michael and Lueckmann, Jan-Matthis and Durkan, Conor and Gon{\c c}alves, Pedro J. and Greenberg, David S. and Macke, Jakob H.},
  title   = {sbi: A toolkit for simulation-based inference},
  journal = {Journal of Open Source Software},
  volume  = {5},
  number  = {52},
  pages   = {2505},
  year    = {2020},
  doi     = {10.21105/joss.02505}
}

@inproceedings{defferrard2016,
  title     = {Convolutional Neural Networks on Graphs with Fast Localized Spectral Filtering},
  author    = {Defferrard, Micha{\"e}l and Bresson, Xavier and Vandergheynst, Pierre},
  booktitle = {Advances in Neural Information Processing Systems},
  year      = {2016},
  eprint    = {1606.09375},
  archivePrefix = {arXiv},
  primaryClass  = {cs.LG}
}

\end{document}